\documentclass[review]{elsarticle}

\usepackage{lineno, hyperref}
\modulolinenumbers[5]

\journal{Journal of Image and Vision Computing}

\begin{document}

\begin{frontmatter}

\title{Transforming the Latent Space of StyleGAN for Real Face Editing}




\author[first_address]{Heyi Li\corref{equal_contribution}}
\cortext[equal_contribution]{Equal contribution}

\author[second_address]{Jinlong Liu\corref{equal_contribution}\corref{corresponding_author}}
\cortext[corresponding_author]{Corresponding author}
\ead{liujinlong@kuaishou.com}

\author[third_address]{Xinyu Zhang}

\author[second_address]{Yunzhi Bai}

\author[second_address]{Huayan Wang}

\author[third_address]{Klaus Mueller}

\address[first_address]{Institute of Remote Sensing, China Academy of Space Technology, Beijing, China}
\address[second_address]{Y-tech, Kuaishou Technology, Beijing, China}
\address[third_address]{Department of Computer Science, Stony Brook University, Stony Brook, NY, USA}

\begin{abstract}
Despite recent advances in semantic manipulation using StyleGAN, semantic editing of real faces remains challenging. The gap between the $W$ space and the $W$+ space demands an undesirable trade-off between reconstruction quality and editing quality. To solve this problem, we propose to expand the latent space by replacing fully-connected layers in the StyleGAN's mapping network with attention-based transformers. This simple and effective technique integrates the aforementioned two spaces and transforms them into one new latent space called $W$++. Our modified StyleGAN maintains the state-of-the-art generation quality of the original StyleGAN with moderately better diversity. But more importantly, the proposed $W$++ space achieves superior performance in both reconstruction quality and editing quality. Despite these significant advantages, our $W$++ space supports existing inversion algorithms and editing methods with only negligible modifications thanks to its structural similarity with the $W/W$+ space. Extensive experiments on the FFHQ dataset prove that our proposed $W$++ space is evidently more preferable than the previous $W/W$+ space for real face editing. The code is publicly available for research purposes at \url{https://github.com/AnonSubm2021/TransStyleGAN}
\end{abstract}


\end{frontmatter}

\nolinenumbers

\section{Introduction}
\label{sec1}

The tremendous success of Generative Adversarial Networks~(GANs)~\cite{NIPS2014_5ca3e9b1} has revolutionized the field of data-driven image generation and has sparked significant research attention. For human face synthesis in particular, the current state-of-the-art architecture StyleGAN~\cite{karras2019style, karras2020analyzing} generates high resolution (i.e. $1024 \times 1024$ pixel), photo-realistic images by first mapping the latent code to layer-wise style code and then feeding it into each convolution layer. Added via adaptive instance normalization~(AdaIN) or its improved technique called weight demodulation, this style code $\mathcal{\textit{S}}$ directly controls image features at various scales. The disentanglement of different attributes (coarse, medium, and fine) exhibited in the intermediate latent space is further explored by numerous follow-up works~\cite{shen2020interpreting, shen2020interfacegan, NEURIPS2020_6fe43269, collins2020editing, shoshan2021gan} to achieve semantically controllable human face generation. However, this manipulation capability is not directly applicable to real faces. 

To mitigate this problem, an “invert then edit” methodology is adopted \cite{shen2020interpreting, shen2020interfacegan, abdal2020image2stylegan++, zhu2020domain, abdal2020styleflow, tewari2020pie, hou2020guidedstyle}. A real image is first projected into the latent space of StyleGAN. Then new latent codes are obtained by performing semantically meaningful edits on inverted latent codes. However, projected latent codes in the $W$ space are not adequate for accurate regeneration of original images. The extended latent space $W$+ has been shown to be more powerful for inversion \cite{abdal2019image2stylegan}. But editing latent codes in the $W$+ space is notoriously ill-posed because they fall out of the semantically meaningful manifold. In contrast, latent codes in the $W$ space do not suffer from this problem and thus favor editing quality. This divide between the $W$ space and the $W$+ space, unfortunately, mandates a compromise between reconstruction accuracy and manipulation naturalness.

\begin{figure*}[t]
    \centering
    \includegraphics[width=\textwidth]{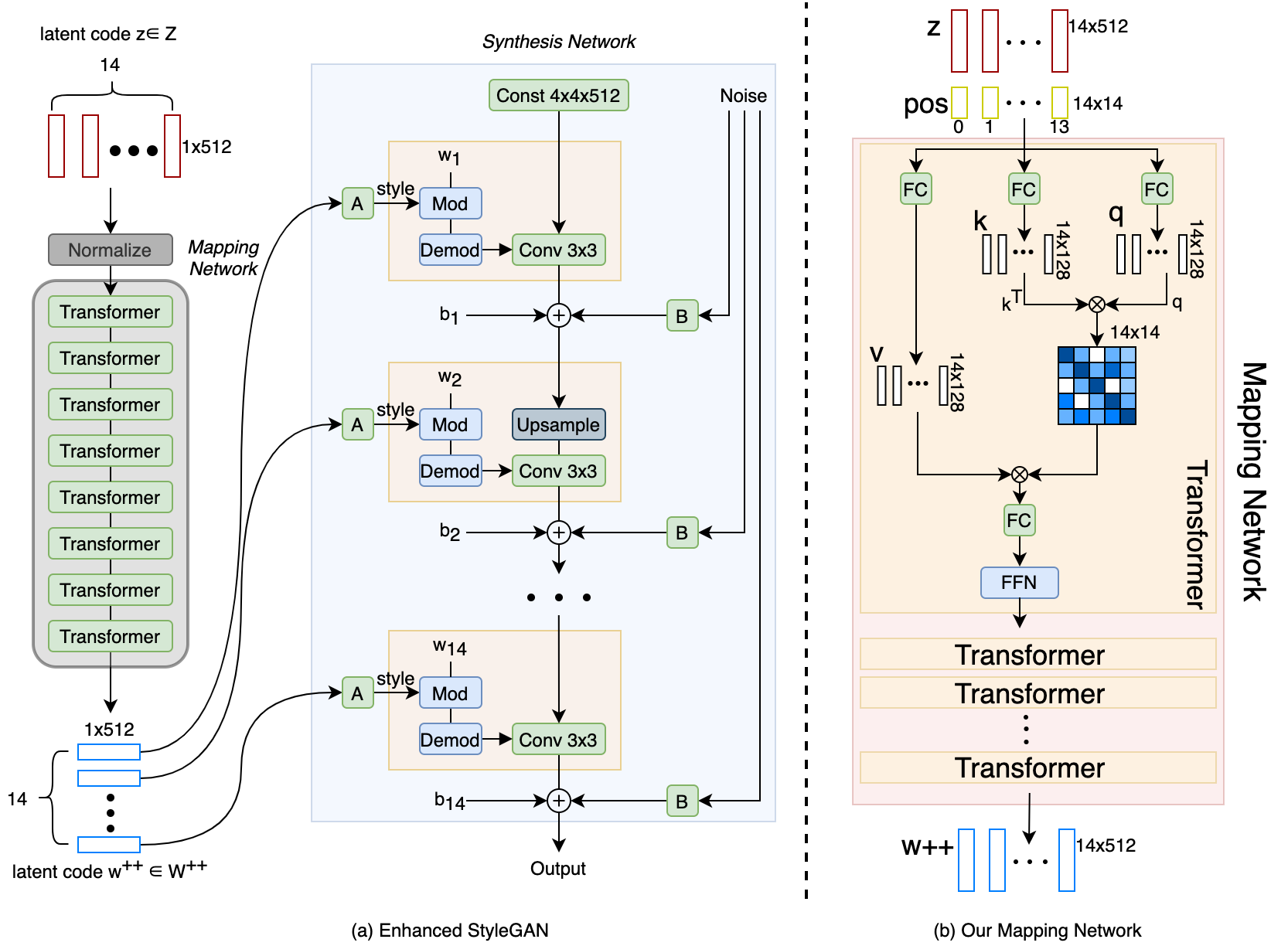}
    \caption{(a) \textbf{Model overview.} While the original StyleGAN~\cite{karras2020analyzing} feeds the same style code to each convolution layer, we redesign the architecture of the StyleGAN mapping network using Transformer so that different style codes are provided to different layers. We illustrate the structure for resolution at $256 \times 256$. (b) An illustration of our proposed mapping network. We omit some standard layers that are not essential for our architecture to not clutter the exposition.}
    \label{fig:fig1}
\end{figure*}

In this paper, we present a straightforward but crucial enhancement to the StyleGAN architecture which expands the dimension of its latent space for better real face editing. We achieve this by replacing the fully-connected layers in the mapping network $M$ with attention-base transformer structures~\cite{NIPS2017_3f5ee243}. This modified mapping network $M'$ maps a random input vector in the $Z$ space to a new intermediate latent space named $W$++. The new $W$++ space has the same dimension as the $W$+ space, which is much larger than the $W$ space and thus achieves a finer reconstruction quality. Unlike latent codes in the $W$+ space, latent codes in the proposed $W$++ space do not fall out of the semantically meaningful manifold after manipulation because they are directly utilized for image synthesis during training. As a result, our $W$++ space enjoys a better editing quality. In addition, our enhanced StyleGAN maintains the state-of-the-art performance of the original StyleGAN with a moderate improvement in terms of generation diversity. Lastly, our proposed $W$++ space readily works with existing inversion algorithms and editing methods requiring only minor adjustments. 

To demonstrate our $W$++ space's outstanding compatibility, we apply a few prevailing methods for inversion and editing, respectively. For the task of inversion, we choose the StyleGAN projector~\cite{karras2020analyzing} to represent optimization-based methods and the pixel2style2pixel~(pSp)~\cite{richardson2020encoding} to illustrate learning-based methods. For the task of editing, we start with the InterfaceGAN~\cite{shen2020interpreting, shen2020interfacegan} for single attribution manipulation. Inspired by the idea of conditional GAN~(cGAN)~\cite{mirza2014conditional}, we then propose a cGAN-based pipeline for attribute editing of real faces. Unlike traditional cGANs, our pipeline~(see Figure~\ref{fig:fig6}) uses attributes extracted by pre-trained models as conditioning information. Another difference is that our discriminator learns to distinguish real and fake images without any conditioning information and thus is same as one in a naive GAN. 

Our main contributions are summarized as follows:

\begin{itemize}

\item We propose to enhance the StyleGAN model by replacing fully-connected layers in its mapping network with attention-based transformers. This augmentation keeps the state-of-the-art generation quality and also moderately improves the generation diversity. 

\item Our proposed latent space $W$++ achieves superior performance in both reconstruction quality and editing quality. 

\item Our new $W$++ space offers excellent compatibility with existing algorithms for both inversion and editing. Only minor changes are needed. 

\end{itemize}

\section{Related Work}
\label{sec2}

In the following, we briefly describe the existing literature on the subject of real face editing. The "invert then edit" methodology has become the de facto standard in this active research field. In the first inversion step, a given image is projected back into the latent space of StyleGAN and the inverted latent code can faithfully reconstruct the input image through the generator. Then new latent codes are obtained by changing the old ones along semantically meaningful directions in the latent space. 

\subsection{GAN Inversion}
\label{sec2.1}

The recent survey by Xia et al.~\cite{xia2021gan} provides an exhaustive overview of GAN inversion algorithms. Existing methods can be classified into three main categories: optimization-based, learning-based, and a hybrid approach. 

Optimization-based algorithms iteratively improve a latent code to minimize the error for a given image. Using an additive ramped-down noise, the original StyleGANv2~\cite{karras2020analyzing} proposes to embed images in the $W$ space which enables better editing at the cost of worse reconstruction. To the contrary, Image2StyleGAN and Image2StyleGAN++~\cite{abdal2019image2stylegan, abdal2020image2stylegan++} embed images into the extended $W$+ space which effectively optimizes a separate style for each scale. This approach sacrifices editing quality for reconstruction quality. To find a better balance, PIE~\cite{tewari2020pie} and StyleGAN2Encoder~\cite{rolux2020s2e} adopt a two-stage encoding process which first embeds an image in the $W$ space and then refines its initial latent code by optimizing in the $W$+ space. 

Learning-based methods~\cite{guan2020collaborative, richardson2020encoding, tov2021designing} aim to train an encoder network which maps an image to the latent space. Compared with optimization-based algorithms, learning-based methods have the advantage of low computation complexity but suffer from inferior reconstruction quality. 

The hybrid type such as~\cite{zhu2020domain} combines the above two techniques where an encoder network is first used to obtain an approximate latent code and then this latent code is improved with optimization. 

\subsection{Latent Space Manipulation}
\label{sec2.2}

Given its approximate linearity, the latent space of StyleGAN has been the primary target for semantic manipulation. Supervised methods find linear directions that correspond to changes in a given binary labeled attribute (such as young \textit{vs.} old) with the supervision of semantic annotations. StyleRig~\cite{tewari2020stylerig} utilizes a pre-trained 3DMM to find the mapping between rigging information and face manipulation. StyleFlow~\cite{abdal2020styleflow} modifies a set of predetermined attributes by learning a transformation between different vectors in the $W$+ space. InterfaceGAN~\cite{shen2020interpreting, shen2020interfacegan} trains linear support vector machines (SVMs) to classify latent codes based on semantic labels and uses the normal vector of each hyperplane as the latent direction of the selected attribute. 

To facilitate attribute manipulation in an unsupervised manner, GANSpace~\cite{NEURIPS2020_6fe43269} performs PCA on the sampled data to find primary directions in the latent space. Conversely, Collins~\textit{et al.}~\cite{collins2020editing} discovers the connection between local semantics and components of latent codes using k-means clustering. Finally, SeFa~\cite{shen2020closed} is a closed-form factorization method which computes interpretable directions without any kinds of training or optimization. 

\subsection{Other Space}
\label{sec2.3}

Most recently, a few concurrent works~\cite{zhu2020improved, liu2020style, wu2020stylespace} have been proposed to address this ``reconstruction-editing'' conundrum by exploring other space options. 

Although having achieved a satisfactory trade-off, \cite{zhu2020improved} fails to eliminate the fundamental conflict caused by performing reconstruction and editing in two different latent spaces. Their proposed $P$ space is transformed from the $W$ space by inverting the last Leaky ReLU layer in the StyleGAN mapping network, while the $P$+ space is extended from the $P$ space by concatenation in a similar way as the $W$+ space being extended from the $W$ space. 

Instead of the latent space, \cite{liu2020style, wu2020stylespace} investigate the style space which is spanned by all possible style vectors. However, as is pointed out in \cite{wu2020stylespace}, the style space achieves worse manipulation naturalness and closer reconstruction when compared with the $W$+ space. This, unfortunately, even aggravates the problem that we are trying to solve. 

\begin{table*}[t]
    \caption{Quantitative comparison of our model and the original StyleGAN on generation quality with the FFHQ dataset. $\uparrow$ indicates that higher is better, and $\downarrow$ that lower is better.}
    \centering
    \begin{tabular}{l|l|l|l|l|l}
    \hline
    Model  & FID $\downarrow$  & Precision $\uparrow$  & Recall $\uparrow$  & Density $\uparrow$  & Coverage $\uparrow$  \\
    \hline\hline
    StyleGANv2  & $4.69$  & $\mathbf{0.670}$  & $0.430$  & $\mathbf{1.278}$  & $0.955$  \\
    \hline
    Ours  & $\mathbf{4.67}$  & $0.662$  & $\mathbf{0.454}$  & $1.261$  & $\mathbf{0.961}$  \\
    \hline
    \end{tabular}
    \label{tab:tab1}
\end{table*}

\section{Method}
\label{sec3}

\begin{figure}[t]
    \centering
    \includegraphics[width=0.8\columnwidth]{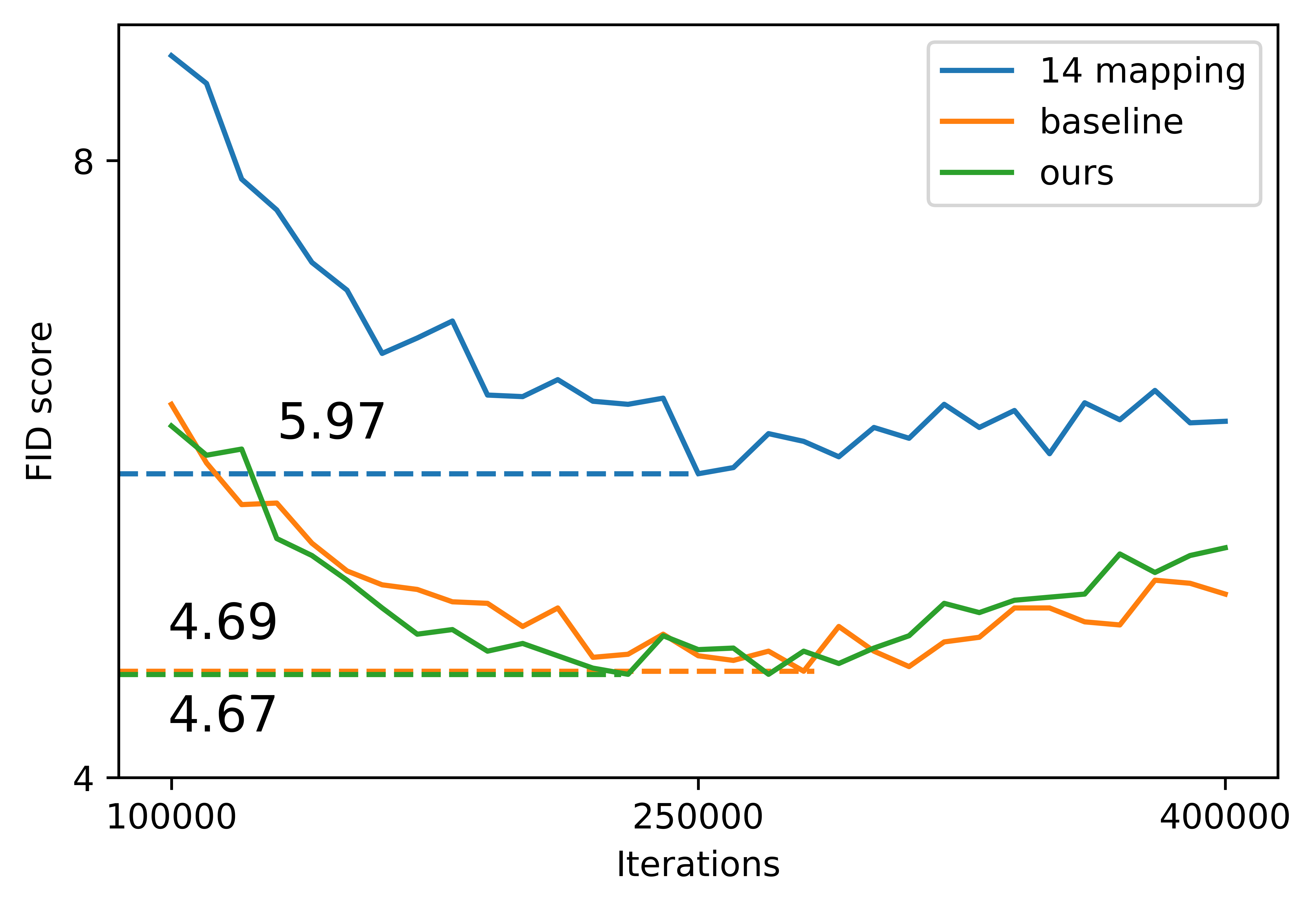}
    \caption{FID score curve for different StyleGAN models at $256 \times 256$ resolution during training. The StyleGAN with $14$ independent mapping networks reaches its optimal score $5.97$ at Iteration $250$K. Our enhanced StyleGAN model achieves the best FID score $4.67$ at Iteration $230$K.}
    \label{fig:fig2}
\end{figure}

\subsection{Motivation}
\label{sec3.1}

The fundamental reason for having to sacrifice either reconstruction quality or editing quality for the other is the fact that these two operations are best suitable in two separate latent spaces. The original StyleGAN uses the same style vector for all varied scales during image generation, which essentially constrains the dimension of the intermediate latent space $W$ to be $1 \times 512$. By traversing the $W$ space, we can find semantically meaningful directions for editing. During inversion, on the other hand, such restriction is lifted which enlarges the dimension of the $W$+ space to be $k \times 512$ ($k$ is the number of style codes). This extra flexibility allows for a more faithful restoration, but it also distorts all editing directions. To resolve this contradiction, we need to expand the input space $Z$ to be of dimension $k \times 512$ and in turn expand the intermediate latent space $W$. But how do we achieve that? (For simplicity and conformity, we fix the image resolution to $256 \times 256$ and thus set the value of $k$ to $14$ for the following analysis.)

One straightforward solution is to expand the dimension of fully-connected layers in the mapping network from $512$ to $7168$ (which is equal to $14 \times 512$). Although the dimension of the intermediate latent space is successively increased in this way, the computation cost also grows quadratically. The estimated FLOPs of a fully-connected layer is $512 \times 512$. And the estimated FLOPs of an expanded fully-connected layer becomes $7168 \times 7168$, which is $196$ times of an original one. And it grows to $324$ times at $1024 \times 1024$ resolution as $k$ reaches $18$. 

Another possible solution is to use an individual mapping network for each layer, which avoids the explosion of computational complexity occurred above. This means $14$ independent mapping networks each creating a different style vector. Since each new mapping network is structurally identical to the original one, this approach limits the computational growth to linear. However, as is shown in Figure~\ref{fig:fig2}, the generation quality indicated by the FID metric deteriorates significantly. The main reason for this degradation is the absence of any correlation among style vectors. In the original StyleGAN, consistency of global features across scales is ensured by feeding the same style vector to all layers in the synthesis network. However, using fully independent mapping networks to compute style vectors completely prohibits such a correlation. 

\subsection{W++ Space}
\label{sec3.2}

An ideal solution to this problem should therefore satisfy the following two conditions: (i) limited growth of computational cost; (ii) some degree of correlation among style vectors. The attention-based transformer structure~\cite{NIPS2017_3f5ee243} fits these two conditions naturally. 

A detailed illustration of our mapping network is provided in Figure~\ref{fig:fig1}(b). We follow the original design of the transformer with a multi-head self-attention and a simple position-wise fully connected feed-forward network. A latent code $z$ is randomly sampled from the input latent space $Z$. This $Z$ space is almost identical to its counterpart in the original StyleGAN except that its dimension is enlarged to $14 \times 512$. Then a fixed 1D position embedding is added to retain the layer index of each style vector~\cite{dosovitskiy2021an}. The query~($q$), the key~($k$), and the value~($v$) all derive from this same latent code $z$ concatenated with position embedding through separate linear transformations. According to the ablation study by Karras et al~\cite{karras2019style} on the depth of the mapping network, $8$ is the optimal choice. Therefore, we implement our mapping network by stacking $8$ transformers. Outputs of this network form our new latent space, $\mathbf{W}$\textbf{++}. 

\paragraph{Computation cost}
In our case, the overall computational complexity of the transformer is dominated by linear transformations instead of the similarity calculations. Each linear transformation is bounded by $O(k d^2)$ where the dimension $d$ is $512$ and the number of style codes $k$ is $14$. We also add a compression ratio $c$ as a trick to reduce the total computation, which lowers the complexity to $O(k (\frac{d}{c})^2)$; in practice, we set the value of $c$ to $4$. In total, our $W$++ space introduces an approximately five-fold increase in computational cost for the mapping network. However, such expansion has limited impact on the whole model given the fact that most of the StyleGAN's computation burden falls upon the synthesis network and the discriminator. 

\paragraph{Style correlation}
The cosine similarity between two linear transformations of the same input, the query and the key, is calculated by the self-attention module. This similarity matrix is then multiplied with the value which is another linear transformation of the input. Therefore, the latent code in our $W$++ space is essentially a weighted sum of the input latent code in the $Z$ space along the $k$ ($=14$) dimension. All coefficients are learned during the training stage. Hence, the mapping network of the original StyleGAN is a special case of ours where correlation between any two elements in the $k$ dimension is rigidly constrained to $1$. Using $14$ independent mapping networks is at the other end of the spectrum because the correlation is then equivalent to $0$. Depending on the input values, our correlation coefficients fall within $[0, 1]$ after training and vary for different pairs. 

As is shown in Figure~\ref{fig:fig1}(a), the output latent code is divided into $14$ different style codes along the $k$ dimensions. Each style code with a dimension of $512$ is then fed to a different layer in the synthesis network at different scale. No changes are made to the original styleGAN architecture to accommodate the proposed $W$++ space. 

\section{Experiments}
\label{sec4}

We evaluate our proposed $W$++ space using the FlickrFaces-HQ (FFHQ) dataset~\cite{karras2019style} from three aspects: generation quality, reconstruction quality, and editing quality. All experiments are performed using PyTorch at $256 \times 256$ resolution with $8$ NVIDIA Tesla V100 GPUs. 

\begin{table*}[t]
    \caption{Quantitative comparison on image reconstruction with StyleGAN projector~\cite{karras2020analyzing} in different latent spaces. Reported results are average values over $400$ images. $\downarrow$ means lower number is better.}
    \centering
    \begin{tabular}{l|l|l|l}
    \hline
    Metrics  & inversion in $W$ space  & inversion in $W$+ space & inversion in $W$++ space  \\
    \hline\hline
    avg Perceptual loss $\downarrow$  &  $0.2079 \pm 0.0389$  &  $0.0855 \pm 0.0177$  &  $0.0937 \pm 0.0179$  \\
    \hline
    avg MSE loss $\downarrow$  &  $0.0776 \pm 0.0512$  &  $0.0217 \pm 0.0179$  &  $0.0255 \pm 0.0183$  \\
    \hline
    \end{tabular}
    \label{tab:tab2}
\end{table*}

\begin{figure*}[t]
    \centering
    \includegraphics[width=\linewidth]{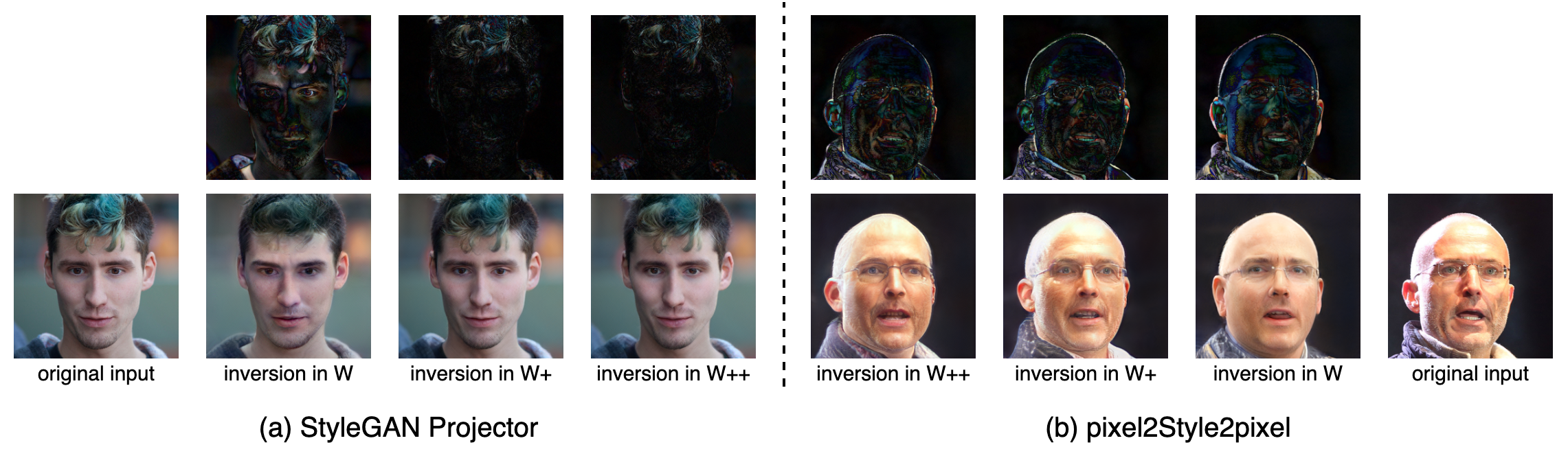}
    \caption{(a) Qualitative comparison on image reconstruction with StyleGAN projector~\cite{karras2020analyzing} in different latent spaces. (b) Qualitative comparison on image reconstruction with pSp encoder~\cite{richardson2020encoding} in different latent spaces. Top row displays differences between inverted images and original images.}
    \label{fig:fig3}
\end{figure*}

\subsection{Image Synthesis}
\label{sec4.1}

We first evaluate our proposed $W$++ space by comparing it with the original StyleGANv2 \footnote{Our work builds upon the PyTorch implementation of StyleGANv2 by rosinality, which is publicly available at \url{https://github.com/rosinality/stylegan2-pytorch}}. We train both models from scratch using exactly the same hyperparameters. The generation quality is measured by Frechet inception distances (FID)~\cite{NIPS2017_8a1d6947} and the diversity is reflected by Recall~\cite{NEURIPS2019_0234c510} and Coverage~\cite{naeem2020reliable}~\footnote{The code is publicaly available at \url{https://github.com/clovaai/generative-evaluation-prdc}}. The results are provided in Table~\ref{tab:tab1}. 

Our modified StyleGAN achieves comparable generation quality than the original StyleGANv2 \footnote{The best FID score announced using rosinality's Pytorch implementation at $256 \times 256$ resolution is $4.5$. While the best FID score we have achieved after multiple runs is $4.69$ at the same resolution. However, this performance gap does not affect our findings because the training codes are identical. }. Furthermore, there is a slight shift from precision to recall as our model achieves higher recall and lower precision. But as is pointed out in \cite{karras2020analyzing}, this is generally desirable because recall can be traded into precision via truncation, whereas the opposite is not true. A similar phenomenon is observed in the density and coverage pair, where density measures fidelity and coverage quantifies diversity. 

In summary, our new StyleGAN maintains the state-of-the-art generation quality of the original StyleGAN. Also, our model gains slightly in terms of diversity thanks to its larger latent space $Z$. 

\subsection{Real Face Inversion}
\label{sec4.2}

In this experiment, we compare inversion in the proposed $W$++ space with both $W$ and $W$+ space. We choose two well-established inversion algorithms, the projector from \cite{karras2020analyzing} as an example for the optimization-based methods, and the pSp~\cite{richardson2020encoding} as an instance for the learning-based methods. Apart from its outstanding reconstruction quality, we also demonstrate that our $W$++ space readily supports existing works. 

\subsubsection{Optimization-Based}
\label{sec4.2.1}

Although initially proposed for inversion in the $W$ space, the StyleGANv2's projector has been extended to support inversion in the $W$+ space as well. Since the $W$++ space shares the same dimension as the $W$+ space, this projector easily works in our new space. Quantitative evaluations between inversion in the $W$ space, the $W$+ space, and our $W$++ space are provided in Table~\ref{tab:tab2}. Figure~\ref{fig:fig3}(a) displays qualitative comparisons for visual inspection.

For quantitative examination, we choose two metrics, the mean square error (MSE) loss and the perceptual loss defined as the LPIPS distance~\cite{zhang2018unreasonable} between target images and restored images. We perform optimization on a total of $400$ images and report the average value for both metrics in all three spaces. Our proposed $W$++ space outperforms the $W$ space by a large margin in both metrics. 

For qualitative examination, the $W$ space clearly performs the worst in reconstruction quality. Due to its low dimension, it simply fails to faithfully restore the identity of the person in the input image. The $W$+ and our $W$++ space, on the other hand, do not suffer from this drawback. Inverted images in these two spaces are almost indistinguishable because they have the same dimensions. These findings are also confirmed by the element-wise difference map shown in the top row of Figure~\ref{fig:fig3}(a). 

Therefore, our proposed $W$++ space clearly outperforms the $W$ space in reconstruction quality while it achieves comparable results with the $W$+ space. 

\begin{table*}[t]
    \caption{Quantitative comparison on image reconstruction with pSp~\cite{richardson2020encoding} in different latent spaces. Reported results are average values over $400$ images. $\downarrow$ means lower number is better.}
    \centering
    \begin{tabular}{l|l|l|l}
    \hline
    Metrics  & inversion in $W$ space  & inversion in $W$+ space & inversion in $W$++ space  \\
    \hline\hline
    avg Perceptual loss $\downarrow$  &  $0.4158 \pm 0.0590$  &  $0.3557 \pm 0.0517$  &  $0.3641 \pm 0.0530$  \\
    \hline
    avg MSE loss $\downarrow$  &  $0.0661 \pm 0.0298$  &  $0.0398 \pm 0.0204$  &  $0.0412 \pm 0.0204$  \\
    \hline
    \end{tabular}
    \label{tab:tab3}
\end{table*}

\subsubsection{Learning-Based}
\label{sec4.2.2}

Our proposed $W$++ space also supports existing learning-based inversion algorithms such as pSp~\cite{richardson2020encoding}~\footnote{The code is publicly available at \url{https://github.com/eladrich/pixel2style2pixel}.} effortlessly. The only modification necessary is to replace the previous StyleGAN model with our enhanced one. We follow the default hyperparameters for training. 

The quantitative and qualitative results are shown in Table~\ref{tab:tab3} and Figure~\ref{fig:fig3}(b) respectively. We adopt the same numerical metrics as above. In accordance with findings in previous works, the reconstruction quality of the learning-based pSp is worse than the reconstruction quality of the optimization-based StyleGAN projector. Again, our proposed $W$++ space accomplishes equivalent superiority with the $W$+ space in reconstruction quality. 

\subsection{Real Face Editing}
\label{sec4.3}

After projecting real images into the latent space, we perform manipulation in the obtained latent codes for semantic editing. Again, we choose two methods to prove the advantage of our proposed $W$++ space in editing quality. One is the well-known InterfaceGAN~\cite{shen2020interpreting, shen2020interfacegan} and the other is a conditional GAN-based editing pipeline which uses pre-trained models as attribute extractors. 

\begin{figure*}[t]
    \centering
    \includegraphics[width=0.65\linewidth]{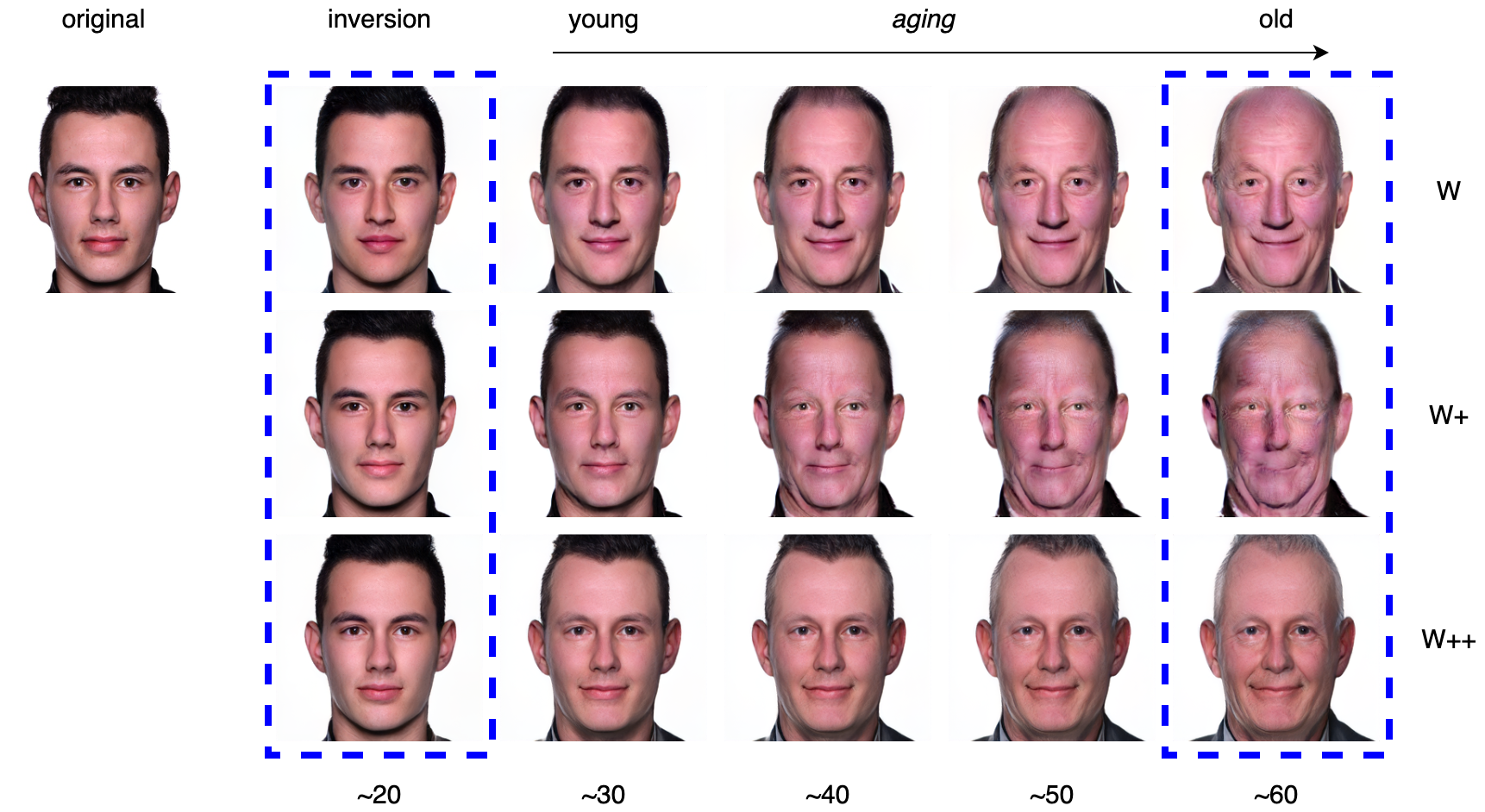}
    \caption{Qualitative comparison of age transformation using InterfaceGAN~\cite{shen2020interpreting, shen2020interfacegan} in different latent spaces. The age label for each image is created using the pre-trained DEX~\cite{Rothe-ICCVW-2015} model. Columns identified by blue boxes are on display in Figure~\ref{fig:fig6}. ``$\sim N$'' denotes ``approximately $N$ years old''. Our results (the bottom row) achieve considerably stronger robustness for long-distance manipulation.}
    \label{fig:fig4}
\end{figure*}

\begin{figure*}[t]
    \centering
    \includegraphics[width=0.65\linewidth]{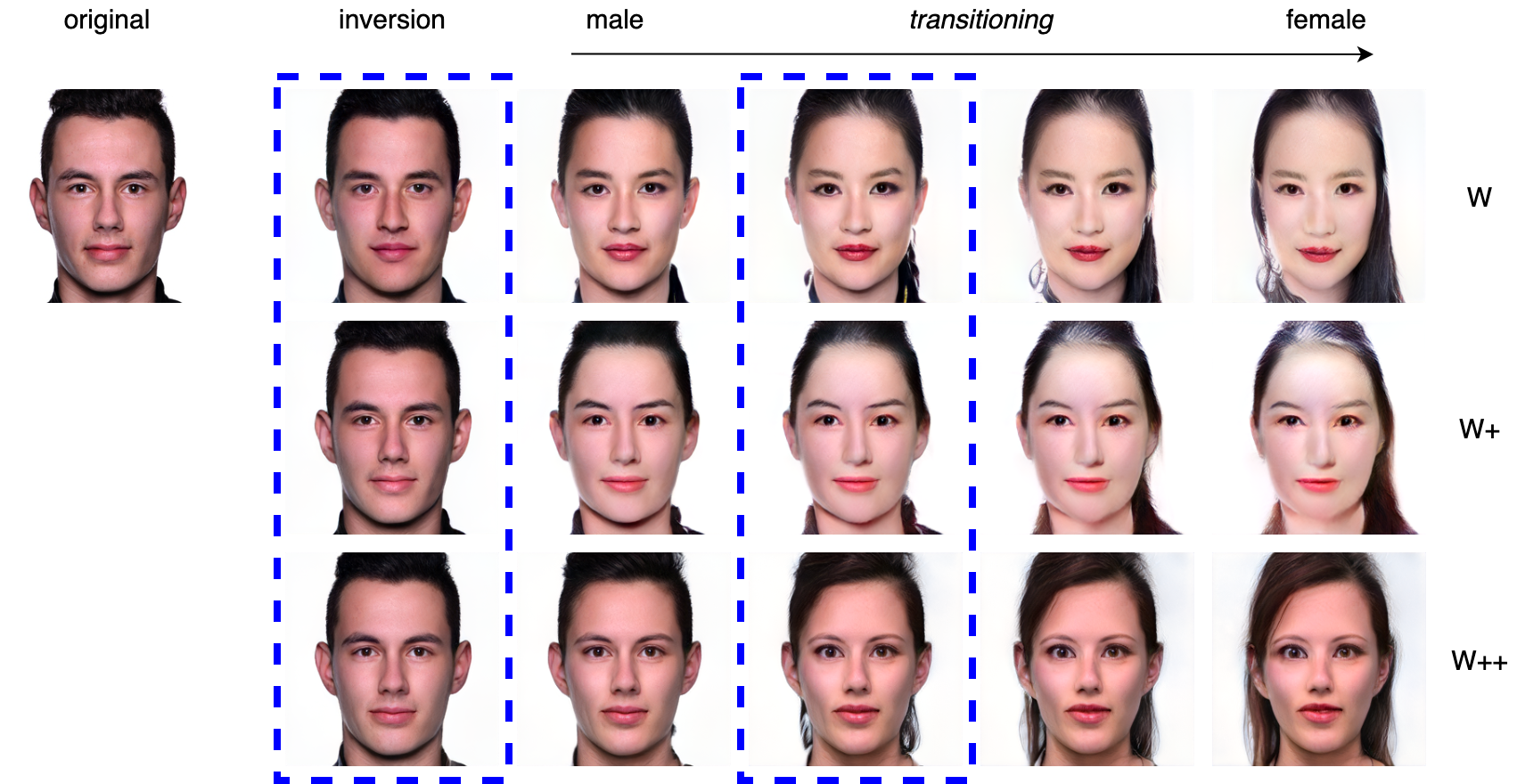}
    \caption{Qualitative comparison of gender transitioning results using InterfaceGAN~\cite{shen2020interpreting, shen2020interfacegan} in different latent spaces. Columns identified by blue boxes are on display in Figure~\ref{fig:fig6}. Our results (the bottom row) achieve considerably stronger robustness for long-distance manipulation.}
    \label{fig:fig5}
\end{figure*}

\subsubsection{InterfaceGAN}
\label{sec4.3.1}

Following the procedure in \cite{shen2020interpreting, shen2020interfacegan}, we train a linear support vector machine~(SVM)~\footnote{The code is publicly available at \url{https://github.com/genforce/interfacegan}.} to find the editing direction which manipulates a real image. $100000$ random latent codes are first generated and then a pre-trained predictor network, such as DEX~\cite{Rothe-ICCVW-2015}~\footnote{The code is publicly available at \url{https://github.com/siriusdemon/pytorch-DEX}.}, is selected to create attribute labels for their corresponding images. Afterwards, these data pairs are used to train the linear SVM. For this demonstration, we use the StyleGANv2's projector to obtain latent codes for real images.

Since the step size plays a crucial role in latent space manipulation, we first conduct an empirical study to align both the visual effect and the classification score of edited images in different spaces. Figure~\ref{fig:fig4} and Figure~\ref{fig:fig5} display the intermediate frames of the entire age transformation and gender transitioning respectively. The first column shows the inverted images and subsequent columns contain images edited to different extents. As the editing distance between the latent code and the classifying boundary keeps increasing, the editing quality of both the $W$ and the $W$+ space starts to deteriorate, especially for editing in the $W$+ space. For gender manipulation in particular (see Figure~\ref{fig:fig5}), changes in the gender direction also transforms the person's race from white to Asian. In contrast, our proposed $W$++ space does not suffer from the ``distance effect'' problem and shows considerably stronger robustness for long-distance manipulation. 

\begin{figure*}[t]
    \centering
    \includegraphics[width=0.6\linewidth]{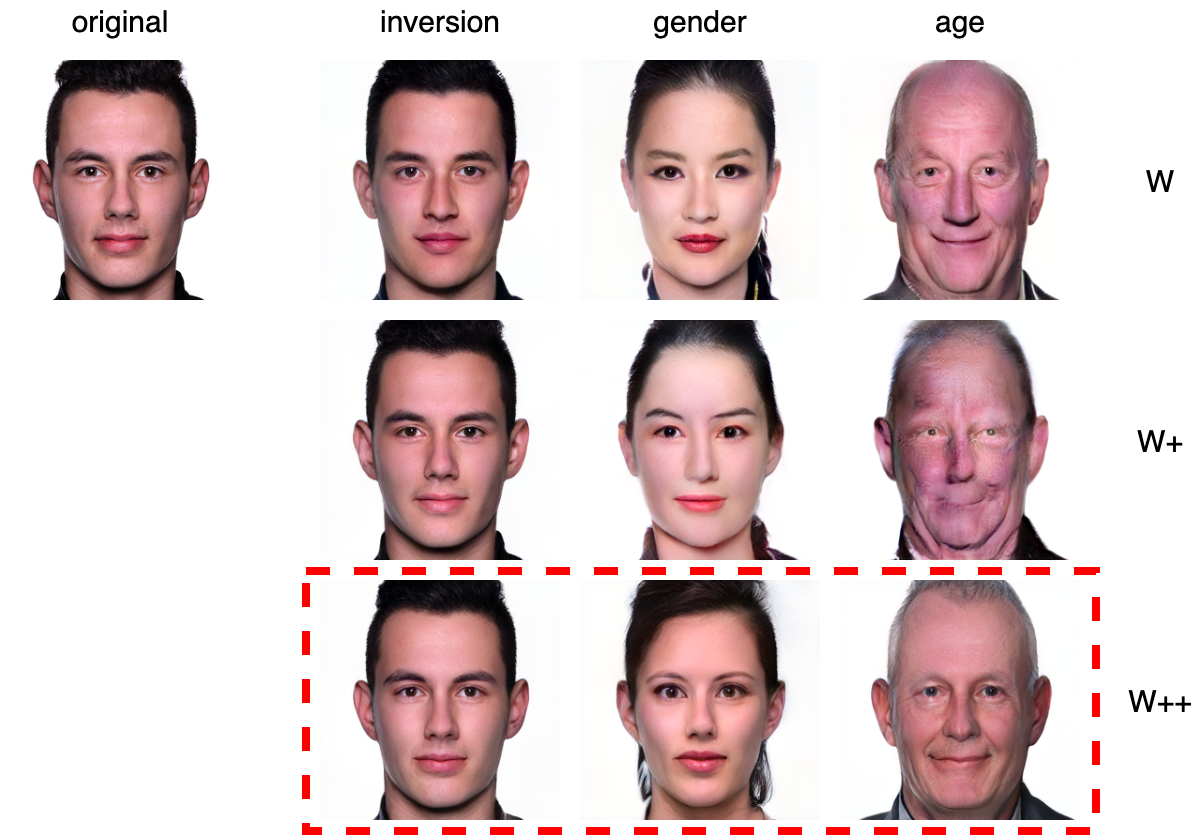}
    \caption{Manipulating real faces with respect to the attribute age in different latent spaces. Given a real image to edit,  we first invert it back to the latent space using StyleGAN projector~\cite{karras2020analyzing} and then manipulate the latent code with InterFaceGAN~\cite{shen2020interpreting, shen2020interfacegan}. Our results (highlighted by the red box) achieve considerably stronger robustness for long-distance manipulation.}
    \label{fig:fig6}
\end{figure*}

The blue boxes indicate the columns we select for each attribute and the final results are shown in Figure~\ref{fig:fig6}. The results in the $W$+ space~(see the middle row) clearly exhibits the worst editing quality in both cases. Although enjoying a better editing quality, the $W$ space suffers from the ``distance effect'' problem~\cite{shen2020interpreting, shen2020interfacegan} where only near-boundary manipulation works well. When the latent code goes further from the boundary, the quality of edited images deteriorates dramatically as manipulating one attribute starts to affect others. In comparison, our proposed $W$++ space~(see the bottom row) achieves an exceedingly more satisfying editing quality. The improvement is extremely obvious in the third column where images are modified to transit gender. In this case, the race of the target person changes from white to Asian when moving the latent code along the gender direction in the $W$ space. Edited images in our $W$++ space faithfully preserve personal identity even for long-distance manipulations. 

Two more examples for age transformation are provided in Figure~\ref{fig:fig10} and Figure~\ref{fig:fig11}. Figure~\ref{fig:fig12} and Figure~\ref{fig:fig13} exhibit two extra examples for gender transitioning.

\subsubsection{cGAN-based Pipeline}
\label{sec4.3.2}

\begin{figure*}[t]
    \centering
    \includegraphics[width=0.8\textwidth]{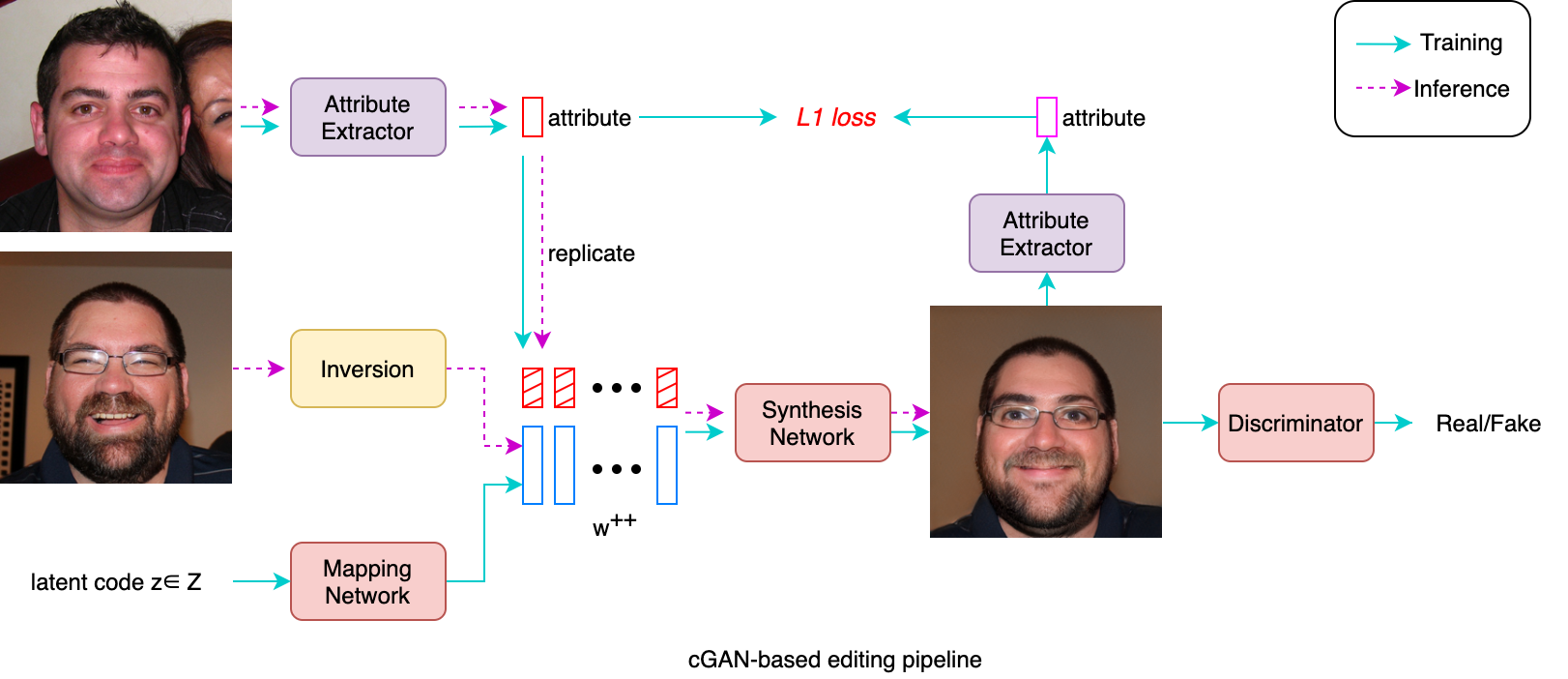}
    \caption{Our conditional GAN-based framework for editing real images. Here we cast one person's facial expression(\textit{eyes open and mouth closed}) to another person.}
    \label{fig:fig7}
\end{figure*}

To further demonstrate the superiority of our proposed $W$++ space, we utilize a conditional GAN-based pipeline for attribute manipulation of real faces. As is shown in Figure~\ref{fig:fig7}, our pipeline adds conditioning information to the StyleGAN generation process. During training, the pre-trained attribute extractor first extracts attribute information from target images. Then this attribute code is concatenated with the latent code in our $W$++ space. Different style codes are concatenated with exactly the same attribute code and are fed to the synthesis network to generate outputs. Afterwards, the same attribute extractor extracts attribute information from generated images. An $L1$ attribute loss between the two attribute codes is minimized as a form of supervision. Unlike traditional cGAN, the discriminator in our case does not take any conditioning information as input. It is only trained to distinguish realistic-looking images from fake ones and thus is identical to the discriminator in StyleGANv2. Also, only the StyleGAN part (the mapping network, the synthesis network, and the discriminator) is trainable since the attribute extractor is fixed. 

For editing real faces, the corresponding latent code for a given input is obtained through inversion. Different attribute classifiers are selected to extract different attributes (such as age, gender, smile, eyes open) from target images. Concatenated with various attribute codes, the synthesis network consequently generates images of the same person but with intended attributes. 

\begin{figure*}[t]
    \centering
    \includegraphics[width=0.8\linewidth]{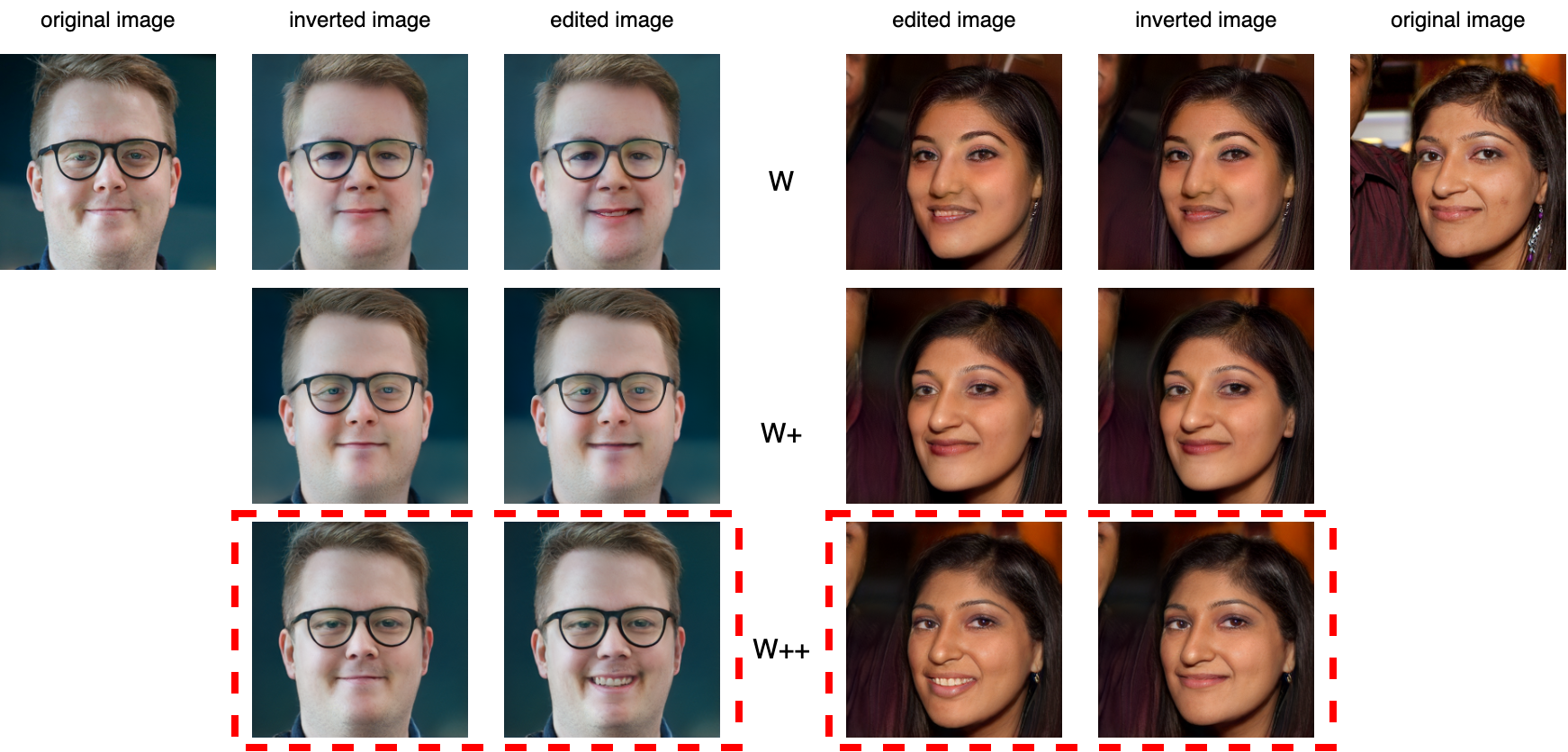}
    \caption{Manipulating real faces with respect to the attribute smile in different latent spaces. Given a real image to edit,  we first invert it back to the latent space using StyleGAN projector~\cite{karras2020analyzing} and then manipulate the latent code with our proposed cGAN-based editing pipeline. Our results (highlighted by the red box) exhibit the most apparent and natural smile expression.}
    \label{fig:fig8}
\end{figure*}

In this experiment, we use the StyleGANv2's projector to invert real images into latent codes. The default hyperparameters are used to train projectors in all three spaces ($W$, $W$+, and $W$++) for a fair comparison. We utilize a pre-trained feature extractor which captures various facial movements essential for depicting a person's smile. The attribute code extracted is a $1D$ vector of $51$ facial coefficients. For editing the smile expression in particular, we extract the attribute code from a single source image and apply it to all input images. The comparison results are shown in Figure~\ref{fig:fig8}. 

Here we show two examples, one female and one male. Inverted images in the $W$ space show the lowest quality, which in turn impairs the quality of edited images and makes the $W$ space ill-suited for real image editing. The $W+$ space achieves excellent reconstruction quality but the worst editing quality. The smile expression after editing is indistinguishable. Our proposed $W++$ space, on the other hand, attains outstanding quality in both reconstruction and editing. The smile expression in the edited images are both apparent and natural. Additionally, our cGAN-based editing pipeline enables exceptional attribute disentanglement. As is shown in the last row of Figure~\ref{fig:fig8}, all irrelevant features remain unaltered. Additional results can be found in Figure~\ref{fig:fig14} and Figure~\ref{fig:fig15}.

\subsection{User Study}
\label{sec4.4}

In addition, we conduct a user study to evaluate our proposed $W++$ space quantitatively. Participants are randomly assigned inverted images and edited images in either the $W$ space or the $W+$ space, paired with corresponding images in the $W++$ space. Then they are asked to select the one of the highest quality according to their own judgements. In this way, we are able to determine which space generates more authentic images to the human eye.


\subsubsection{Details of the User Study}
\label{sec4.4.1}

The user study has $4$ tasks, namely \textit{inversion}, \textit{aging}, \textit{smiling} and \textit{transition}. In each task, the original image is shown to the left along with two generated images to the right. One of the two images is in the proposed $W++$ space, while the other one is randomly picked either in the $W$ space or the $W+$ space. The sequence of these two images shown is shuffled so that the image in the $W++$ space could either be in the second column or the third column. More importantly, all tasks in our study follow the double-blind procedure to guard against any potential bias. 


\textbf{Inversion:} In this task, the goal is to reconstruct an image from a latent code. $3$ sets of images are randomly picked out of a total number of $12$. The participant is then asked to select one image per set which looks more identical to the original image among the two.


\textbf{Aging:} In this task, the goal is to edit a portrait so that the person looks like at $60$ years old. $3$ randomly selected sets of images are shown. And the participant is asked to select the one more realistic per set. 


\textbf{Smiling:} In this task, the goal is to edit a portrait so that the person puts on a smiling face. $2$ sets of images are randomly picked out of a total number of $6$. The participant is then asked to select the more genuine one per set among the two images shown.


\textbf{Transition:} In this task, the goal is to edit an image so that the person is transited to the opposite gender. $2$ randomly selected sets of images are shown. And the participant is asked to select the one more authentic per set.


We recruit a total number of $41$ users with diverse backgrounds. $21$ of them are self-identified as male, while the rest is self-identified as female. There are $12$ participants who are younger than $25$, $19$ between age $25$ and $30$, $7$ between age $30$ and $35$, and 3 above the age of $35$. $11$ people has some experience in the field of computer vision, while the others do not.


\subsubsection{Result Analysis}
\label{sec4.4.2}

\begin{figure*}[t]
    \centering
    \includegraphics[width=0.8\linewidth]{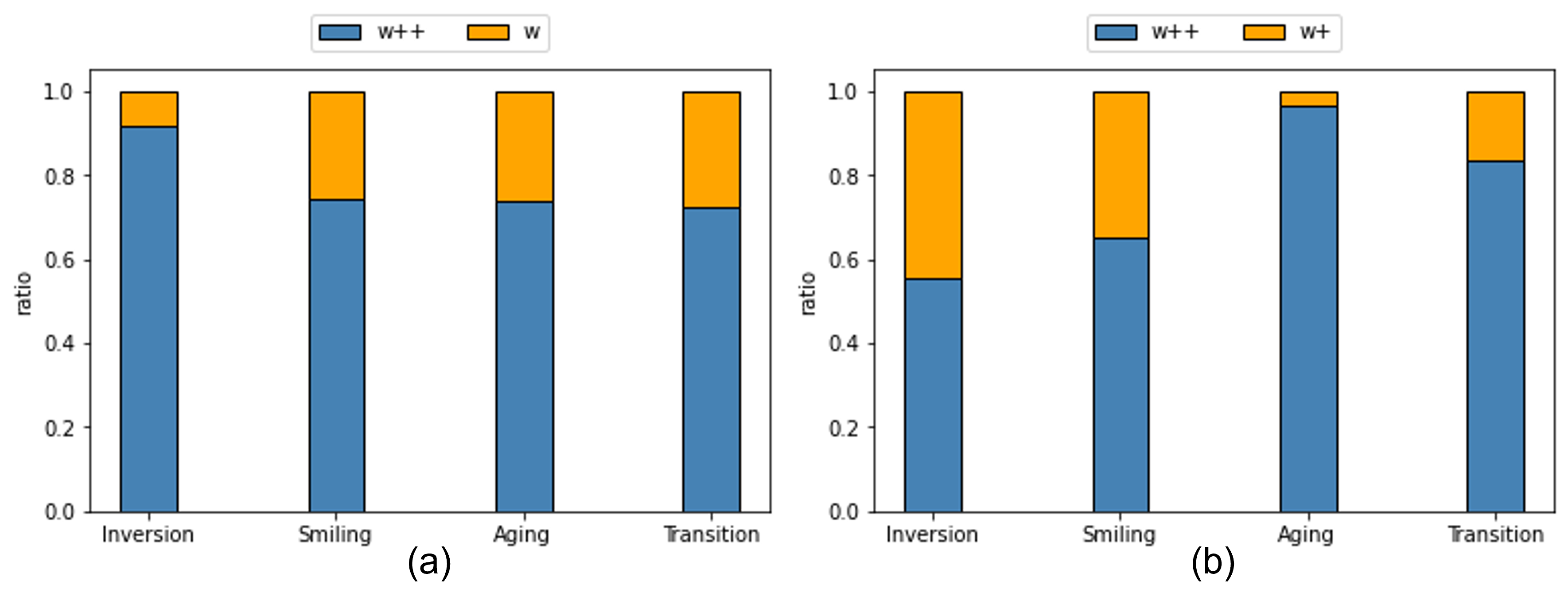}
    \caption{Stacked bar charts of the preference ratio (a) the $W++$ space \textit{vs.} the $W$ space (b) the $W++$ space \textit{vs.} the $W+$ space}
    \label{fig:fig9}
\end{figure*}

\begin{figure*}[t]
    \centering
    \includegraphics[width=0.8\linewidth]{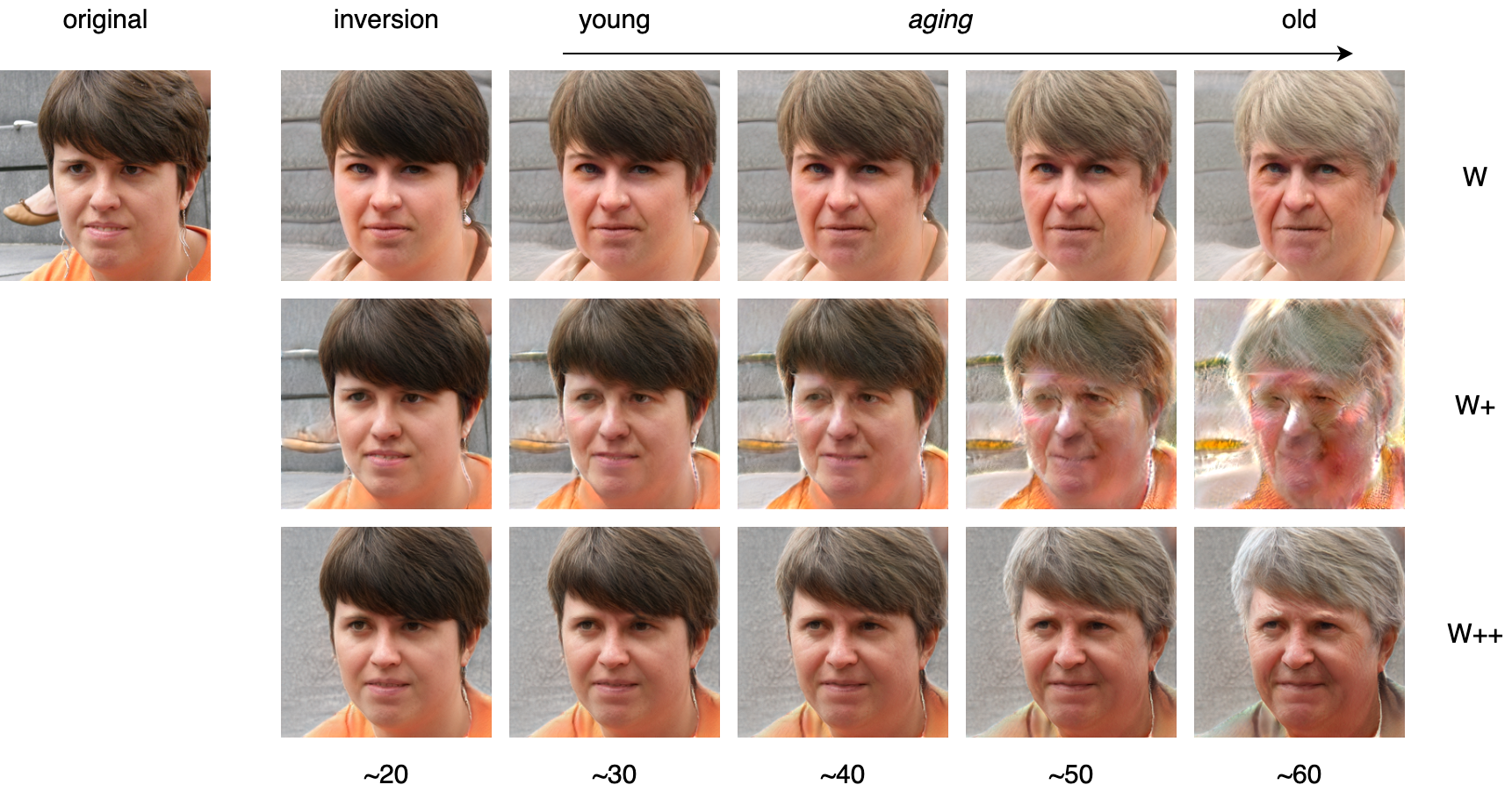}
    \caption{Qualitative comparison of age transformation using InterfaceGAN~\cite{shen2020interpreting, shen2020interfacegan} in different latent spaces. The age label for each image is created using the pre-trained DEX~\cite{Rothe-ICCVW-2015} model. ``$\sim N$'' denotes ``approximately $N$ years old''. Our results (the bottom row) achieve considerably stronger robustness for long-distance manipulation.}
    \label{fig:fig10}
\end{figure*}

\begin{figure*}[t]
    \centering
    \includegraphics[width=0.8\linewidth]{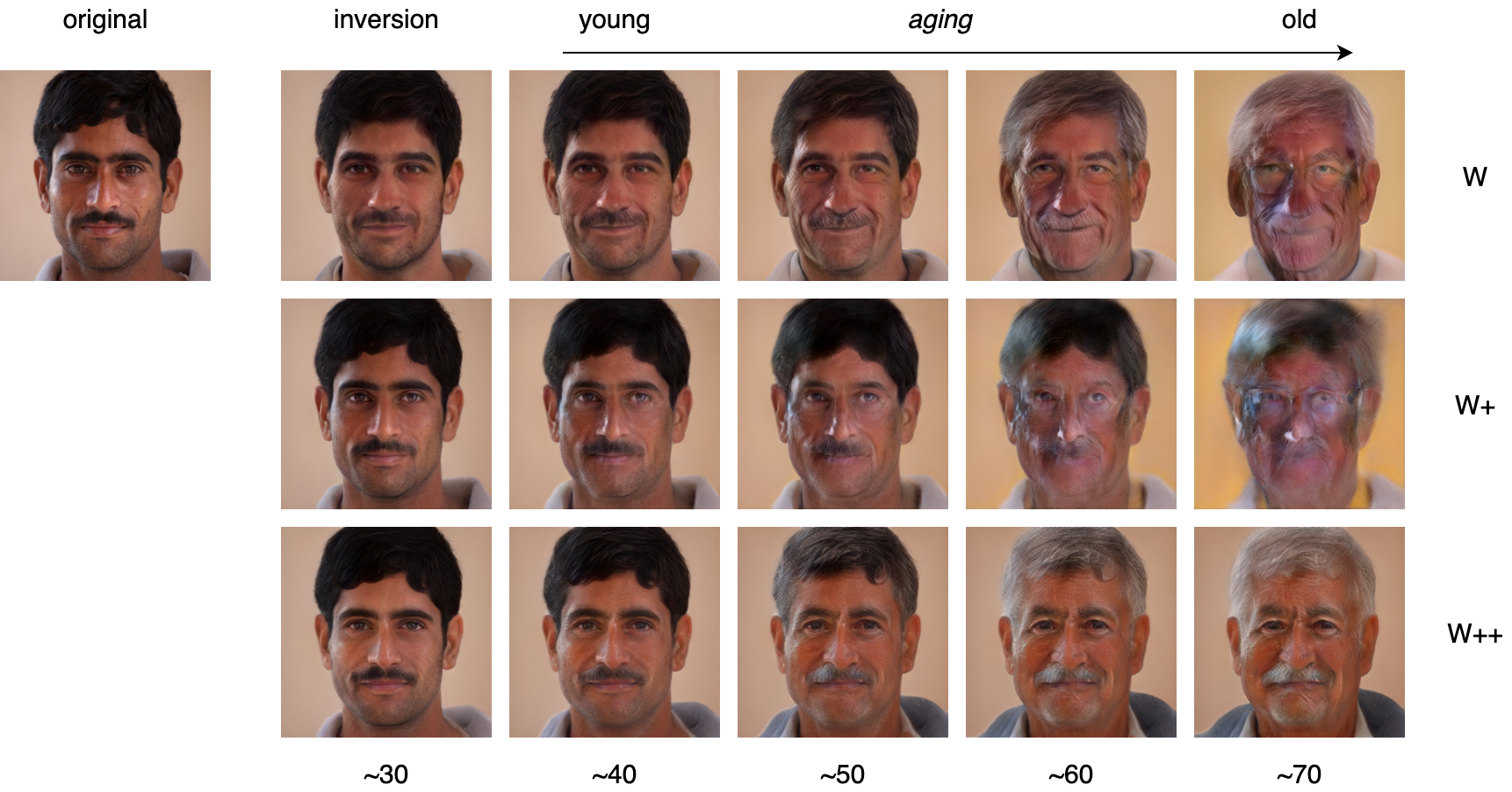}
    \caption{Qualitative comparison of age transformation using InterfaceGAN~\cite{shen2020interpreting, shen2020interfacegan} in different latent spaces. The age label for each image is created using the pre-trained DEX~\cite{Rothe-ICCVW-2015} model. ``$\sim N$'' denotes ``approximately $N$ years old''. Our results (the bottom row) achieve considerably stronger robustness for long-distance manipulation.}
    \label{fig:fig11}
\end{figure*}

\begin{figure*}[t]
    \centering
    \includegraphics[width=0.8\linewidth]{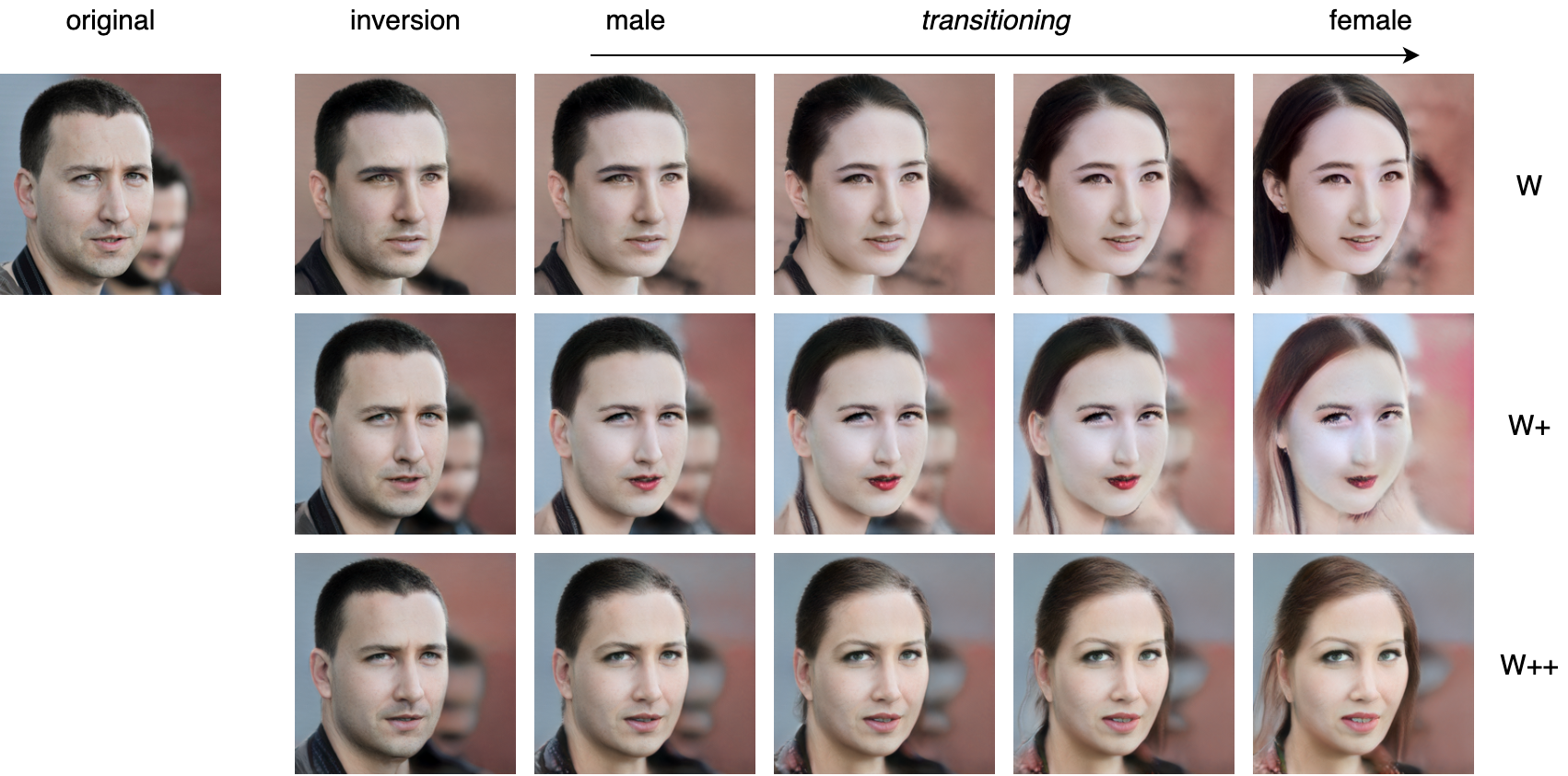}
    \caption{Qualitative comparison of gender transitioning results using InterfaceGAN~\cite{shen2020interpreting, shen2020interfacegan} in different latent spaces. Our results (the bottom row) achieve considerably stronger robustness for long-distance manipulation.}
    \label{fig:fig12}
\end{figure*}

\begin{figure*}[t]
    \centering
    \includegraphics[width=0.8\linewidth]{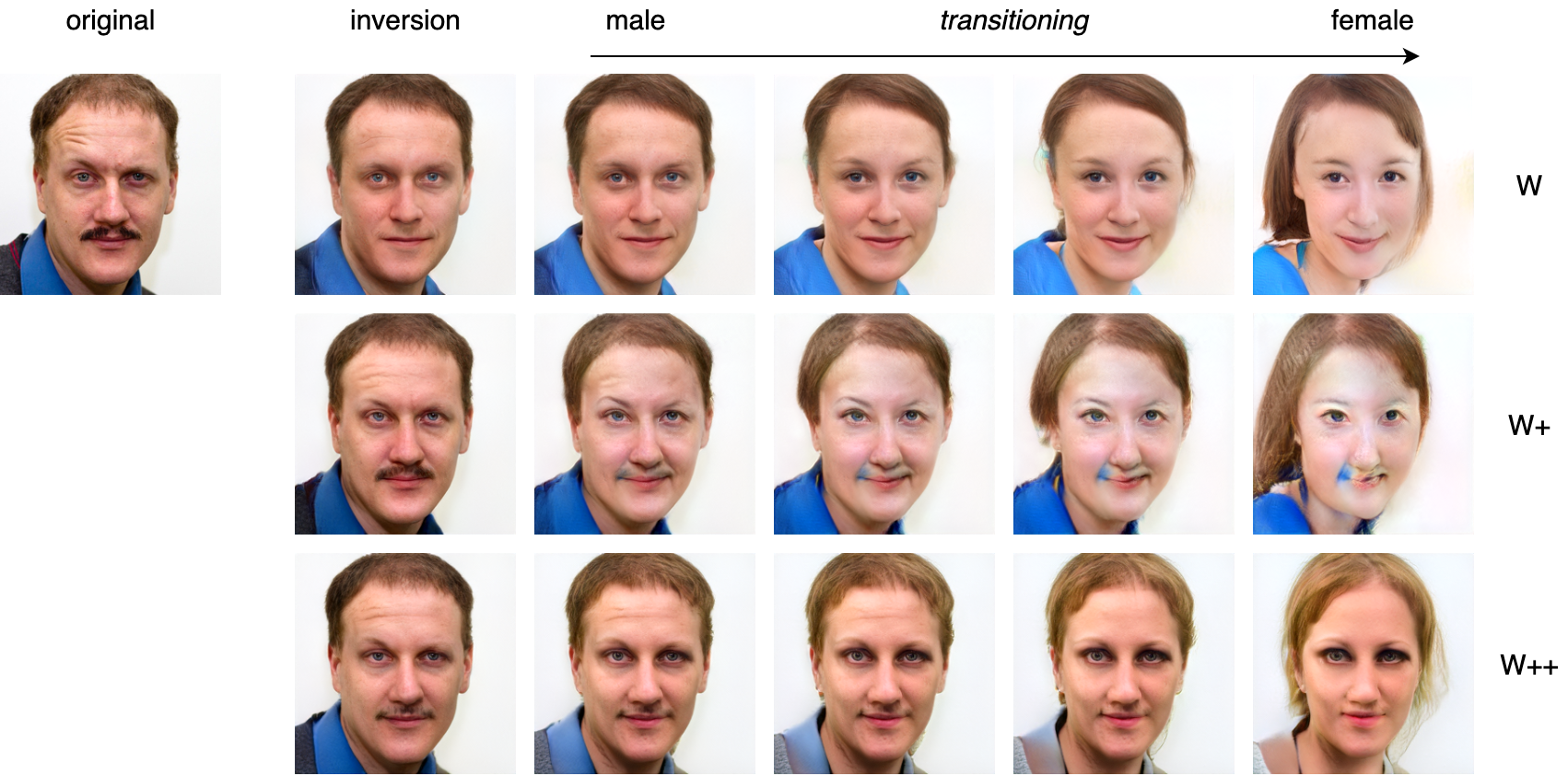}
    \caption{Qualitative comparison of gender transitioning results using InterfaceGAN~\cite{shen2020interpreting, shen2020interfacegan} in different latent spaces. Our results (the bottom row) achieve considerably stronger robustness for long-distance manipulation.}
    \label{fig:fig13}
\end{figure*}

\begin{figure*}[t]
    \centering
    \includegraphics[width=0.8\linewidth]{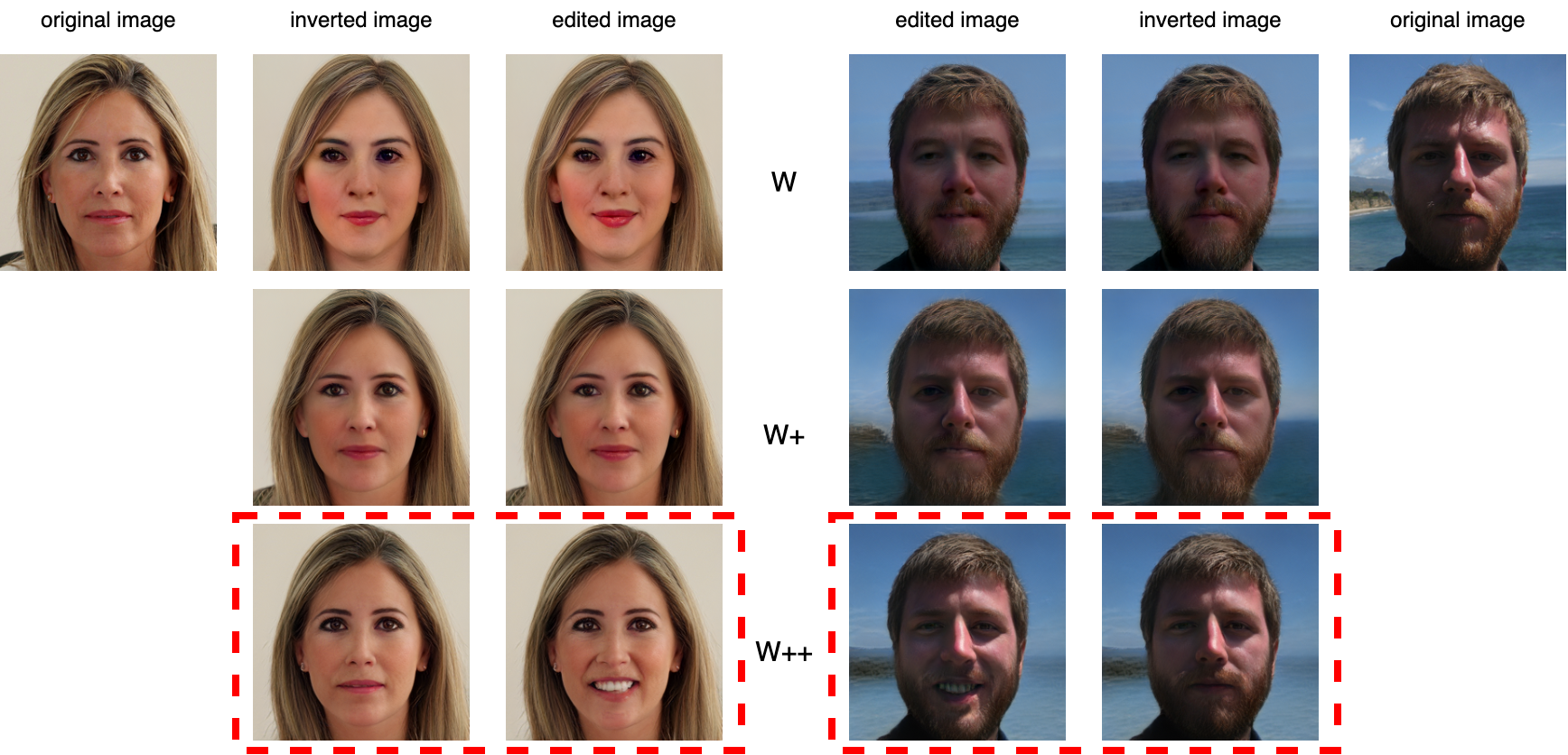}
    \caption{Manipulating real faces with respect to the attribute smile in different latent spaces. Given a real image to edit,  we first invert it back to the latent space using StyleGAN projector~\cite{karras2020analyzing} and then manipulate the latent code with our proposed cGAN-based editing pipeline. Our results (highlighted by the red box) exhibit the most apparent and natural smile expression.}
    \label{fig:fig14}
\end{figure*}

\begin{figure*}[t]
    \centering
    \includegraphics[width=0.8\linewidth]{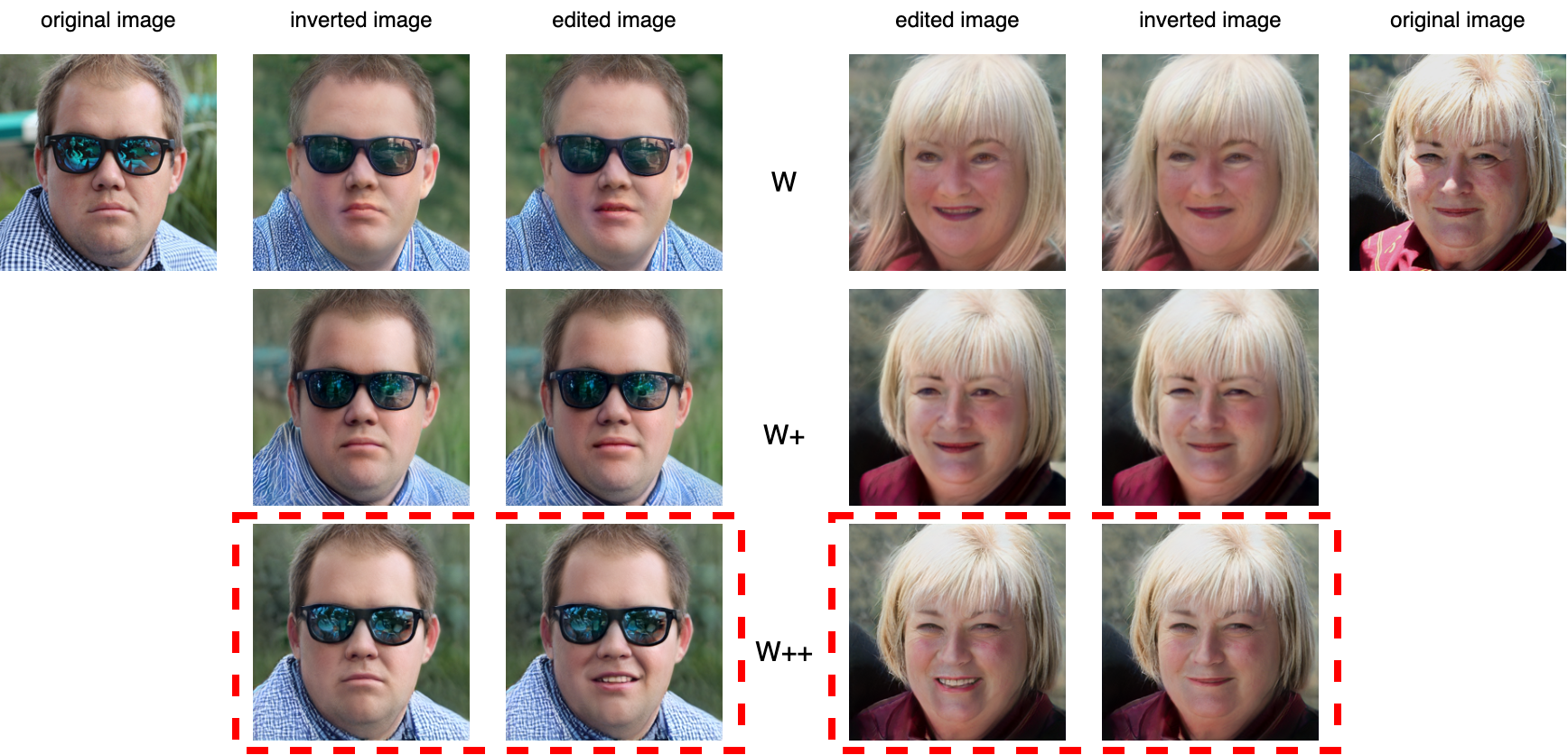}
    \caption{Manipulating real faces with respect to the attribute smile in different latent spaces. Given a real image to edit,  we first invert it back to the latent space using StyleGAN projector~\cite{karras2020analyzing} and then manipulate the latent code with our proposed cGAN-based editing pipeline. Our results (highlighted by the red box) exhibit the most apparent and natural smile expression.}
    \label{fig:fig15}
\end{figure*}

We designate the $W$ space and the $W+$ space as $2$ independent control groups and compare the result of the $W++$ space with each of them respectively. As is shown in Figure~\ref{fig:fig9}, our proposed $W++$ space is overwhelmingly more preferable among participants than either the $W$ space or the $W+$ space in most tasks. For example, $91.7$ percent of users conclude that the reconstruction quality of our $W++$ space is better than the $W$ space(according to Figure~\ref{fig:fig9}(a)) and $96.3$ percent of participants believe that the editing quality for aging in the proposed $W++$ space is better than the $W+$ space(according to Figure~\ref{fig:fig9}(b)). Figure~\ref{fig:fig9}(b) also shows that $55.6$ percent of human subjects prefer the inverted images in our $W++$ space to those in the $W+$ space, which resonates with the fact that these two spaces achieve comparable perceptual loss and MSE loss values. 


To evaluate the statistical significance, we supplement our user study with a two-way ANOVA test and the result is shown in Table~\ref{tab:tab4}. The $p$ values for the models in both tables are less than the threshold of $0.001$ which indicates that the difference of $W++$ and the other model is highly statistically significant rather than caused by random factors. Therefore, our proposed $W++$ space is superior to both the $W$ space and the $W+$ space with highly statistical significance. Furthermore, the $p$ values for the tasks in both tables are more than the threshold of $0.05$. This implies that our results are not attributable to the choice of tasks. As a result, the superiority of our proposed $W++$ space is universal across all tasks. 


\begin{table*}[t]
    \caption{Two-way ANOVA analysis for (a) the $W++$ space \textit{vs.} the $W$ space (b) the $W++$ space \textit{vs.} the $W$ space to analyze the statistical significance.}
    \centering
    \begin{tabular}{l|l|l|l|l|l}
    \hline
    Table (a)  & SS  & df  & MS  & F  & P  \\
    \hline\hline
    Models  & $53.33$  & $1$  & $53.33$  & $127.59$  & $\mathbf{<0.0001}$  \\
    \hline
    Tasks  & $2.24$  & $3$  & $0.75$  & $1.79$  & $0.1494$  \\
    \hline
    Models*Tasks & $6.68$ & $3$ & $2.23$ & $5.33$ & $0.0014$\\
    \hline
    Error & $109.51$ & $262$ & $0.42$ & - & -\\
    \hline
    Total & $171.76$ & $269$ & - & - &-\\
    \hline
    \end{tabular}\\
    
    \begin{tabular}{l|l|l|l|l|l}
    \hline
    Table (b)  & SS  & df  & MS  & F  & P  \\
    \hline\hline
    Models  & 36.11  & 1  & 36.11  & 99.77  & $\mathbf{<0.0001}$  \\
    \hline
    Tasks  & 1.76  & 3  & 0.59  & 1.62  & 0.1851  \\
    \hline
    Models*Tasks & 15.35 & 3 & 5.12 & 14.14 & $<0.0001$\\
    \hline
    Error & 93.38 & 258 & 0.36 & - & -\\
    \hline
    Total & 146.6 & 265 & - & - &-\\
    \hline
    \end{tabular}\\
    
    \label{tab:tab4}
\end{table*}

\section{Conclusion}
\label{sec5}

In this work, we propose to upgrade the StyleGAN architecture by replacing its mapping network with $8$ attention-based transformers. This modification transforms its original latent space to a new latent space called $W$++. Our StyleGAN model retains the state-of-the-art generation quality and moderately improves generation diversity. However, unlike the previous $W$ or $W$+ spaces, our proposed $W$++ space achieves superior performance in both reconstruction quality and editing quality. Additionally, it supports existing inversion algorithms and editing methods with only minor adjustments needed. Experiments using FFHQ dataset clearly demonstrate the merits of our method. 

Our work has some limitations that we leave to future work. Although our proposed $W$++ space has achieved excellent reconstruction quality by expanding the original latent space, the inverted image still looks a little different from the input real image. This dissimilarity adversely impacts the editing quality of real images. In the future, we would like to close this gap.

\section*{Acknowledgments}

We would like to show our gratitude to Jun Fu for advice on the design of figures; Jiayi Liu, Shen Wang, and Zhihang Li for paper reviews. 

This research was partially supported by NSF grants IIS 1527200 and 1941613.


\bibliography{mybibfile}

\end{document}